\documentclass{article}
\usepackage{spconf,amsmath,graphicx}
\usepackage{bm}
\usepackage{url}

\title{Improving robustness to out-of-distribution data \\by frequency-based augmentation}
\name{Koki Mukai, Soichiro Kumano, and Toshihiko Yamasaki}
\address{The University of Tokyo}
\begin{document}
%
\maketitle
\begin{abstract}
Although Convolutional Neural Networks (CNNs) have high accuracy in image recognition, they are vulnerable to adversarial examples and out-of-distribution data, and the difference from human recognition has been pointed out. In order to improve the robustness against out-of-distribution data, we present a frequency-based data augmentation technique that replaces the frequency components with other images of the same class. When the training data are CIFAR10 and the out-of-distribution data are SVHN, the Area Under Receiver Operating Characteristic (AUROC) curve of the model trained with the proposed method increases from 89.22\% to 98.15\%, and further increased to 98.59\% when combined with another data augmentation method. Furthermore, we experimentally demonstrate that the robust model for out-of-distribution data uses a lot of high-frequency components of the image.

\end{abstract}
\section{Introduction}
\label{sec:intro}
In recent years, the accuracy of image recognition performance has been improving. In particular, the ImageNet Large Scale Visual Recognition Challenge (ILSVRC)~\cite{ILSVRC} in 2012, a competition for image recognition accuracy, saw a dramatic improvement in accuracy with the introduction of AlexNet~\cite{Alex2012}, which uses Convolutional Neural Networks (CNNs). However, the vulnerability towards adversarial examples~\cite{AE,AE1,AE2} and fooling images~\cite{Fooling}, overconfidence to out-of-distribution images have also been reported~\cite{OOD_0,OOD_1,OOD_2}. The generalization performance of such CNNs has been studied in relation to the frequency of the input images~\cite{Freq, Freq2, Amplitude}.

Wang et al.~\cite{Freq} argued that CNNs improve accuracy by using regions that are meaningful to humans and those with high-frequencies that cannot be perceived by humans. For this reason, they argued that images such as adversarial examples, which have relatively noisy high-frequency components, have a difference in recognition from humans. They also pointed out that there is a trade-off between the robustness to adversarial examples and the accuracy to normal images, and raised a question about the focusing on accuracy alone. In addition, Chen et al.~\cite{Amplitude} pointed out that the difference between humans and CNNs  is that CNNs are sensitive to the amplitude component of the image, while humans are sensitive to the phase component. They also pointed out that adversarial examples and other examples show that the recognition of CNNs and humans are different because many changes are made to the amplitude component of an image. Based on their assumptions, they proposed an augmentation method, Amplitude-Phase Recombination (APR), to let CNNs pay more attention to the phase of the images.

For out-of-distribution detection, existing studies have proposed dedicated classifiers or attempted to improve the detection accuracy for pre-trained models in various ways~\cite{OOD_0,OOD_1,OOD_2,OOD_3}. While these methods are effective, it would be beneficial if the robustness of the model itself to out-of-distribution data could be improved. In addition, existing data augmentation methods mainly focus on enhancing the accuracy, which may in turn sacrifice robustness. In this study, we propose a frequency-based data augmentation that improves robustness of the model to the out-of-distribution data in a confidence-based out-of-distribution detection method~\cite{OOD_1}. The contributions of our research are as follows.
\begin{itemize}
    \item We propose a new frequency-domain data augmentation technique that enhances the robustness of the model to out-of-distribution data. Since it is a data augmentation method, it can be combined with other methods.
    \item We show that models with high robustness to out-of-distribution data pay more attention to high-frequency components of the images.
\end{itemize}

\section{Proposed method}
\label{sec:method}
\begin{figure*}[t]
    \begin{center}
    \includegraphics[width=0.60\linewidth]{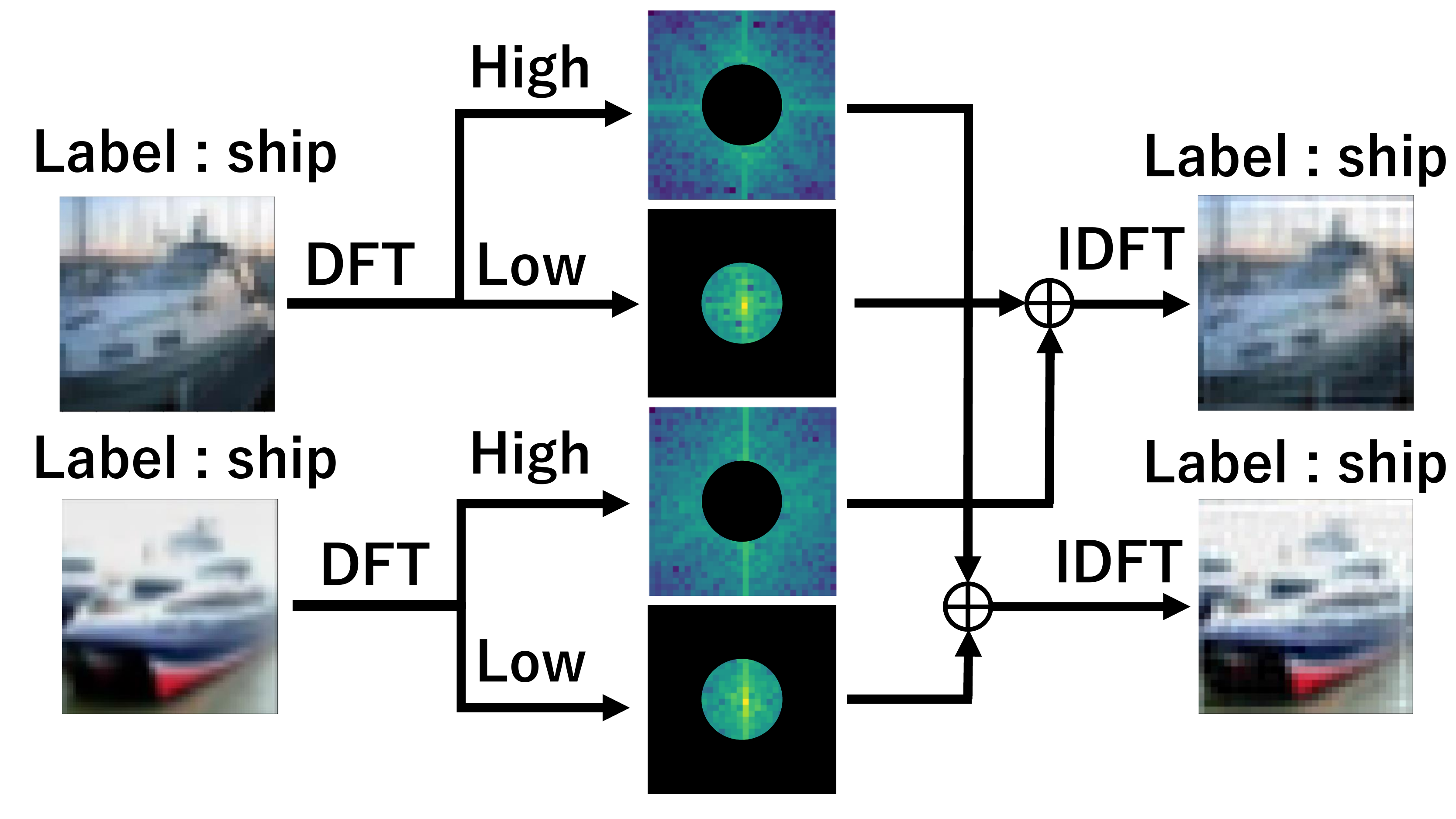}
    \end{center}
      \caption{The flow of RFC. Images taken from the same class are used and their frequency components are swapped.}
    \label{fig:rep}
\end{figure*}
To improve the robustness of CNNs, we propose a frequency-based data augmentation method, in which images are decomposed into low-frequency and high-frequency components and they are swapped with those of other images of the same class (here after, we call this procedure as Replacement of Frequency Component, RFC). While retaining the low and high-frequency information of each class, novel (augmented) images of each class can be generated. The flow of this data augmentation is shown in Figure~\ref{fig:rep} and the detailed procedure is described below.

Let $\bm{x}$ be an image and let us denote its frequency component $\bm{z}$ as $\bm{z} = \mathcal{F}(\bm{x})$ using the Discrete Fourier Transform (DFT) $\mathcal{F}(\cdot)$. We obtain the low-frequency and the high-frequency components of $\bm{z}$, $\bm{z}_l$ and $\bm{z}_h$, respectively, by using the mask matrices to pass the low-frequency and high-frequency components $M_l$ and $M_h$, respectively. With $(c_i, c_j)$ as the indices of the frequency center (DC component) of $M$ and using a radius $r$, $M_l$ and $M_h$ can be described as:
\begin{eqnarray}
      M_l(i,j) = \left\{
        \begin{array}{ll}
          1 & \left( \sqrt{ (i - c_i)^2 + (j - c_j^2) } < r\right) \\
          0 & \left( \sqrt{ (i - c_i)^2 + (j - c_j^2) } \geq r \right) \\
        \end{array}
        \right.,
        \\
        M_h(i,j) = \left\{
          \begin{array}{ll}
            0 & \left( \sqrt{ (i - c_i)^2 + (j - c_j^2) } < r \right) \\
            1 & \left( \sqrt{ (i - c_i)^2 + (j - c_j^2) } \geq r\right)
          \end{array}
          \right. .
    \label{eq:mask}
\end{eqnarray}
Then, $\bm{z}_l$ and $\bm{z}_h$ are defined as:
\begin{eqnarray}
    \bm{z}_l = M_l \otimes \bm{z},\\
    \bm{z}_h = M_h \otimes \bm{z}.
\end{eqnarray}
Here, $\otimes$ is the product of each element. In the same way, let $\bm{z_l'}$ and $\bm{z_h'}$ be respectively the low and high-frequency components of $\bm{x'}$, a randomly selected image from the same class as $\bm{x}$. The frequency-component  swapped images, $\bm{x}_{\mathrm{mix}}$ and $\bm{x}'_{\mathrm{mix}}$, using the two images, $\bm{x}$ and $\bm{x'}$,  are obtained using the inverse Discrete Fourier transform (IDFT) $\mathcal{F}^{-1}(\cdot)$ as:
\begin{eqnarray}
    \bm{x}_{\mathrm{mix}} = \mathcal{F}^{-1}(\bm{z_l}) + \mathcal{F}^{-1}(\bm{z_h'}), \\
    \bm{x}'_{\mathrm{mix}} = \mathcal{F}^{-1}(\bm{z_l'}) + \mathcal{F}^{-1}(\bm{z_h}).
    \label{eq:freq_sys}
\end{eqnarray}
Then, $\bm{x}_{\mathrm{mix}}$ and $\bm{x}'_{\mathrm{mix}}$ are added to the training data.

\section{Experiments}
\label{sec:exp}
\subsection{Experimental setup}
\label{ssec:setup}
In our experiments, the model is ResNet18~\cite{ResNet}, the dataset is CIFAR10~\cite{CIFAR10}, and the optimization method is Stochastic Gradient Descent (SGD), the learning rate started at 0.1, was multiplied by 0.2 at the 60th, 120th, 160th, and 190th epochs, and was continued until the 200th epoch. We use the basic data augmentation of RandomCrop and RandomHorizontalFlip as a baseline. We also employed CutMix~\cite{CutMix}, mixup~\cite{Mixup}, APR~\cite{Amplitude}, RFC (proposed), and RFC+APR for comparison.

We used SVHN~\cite{SVHN}, LSUN~\cite{LSUN}, ImageNet~\cite{ImageNet}, and CIFAR100~\cite{CIFAR10} as out-of-distribution data. For out-of-ditribution detection, we use the Hendrycks's method~\cite{OOD_1}. It uses confidence scores of the model's output to distinguish whether the input image is in-distribution or out-of-distribution. By using True Positive Rate (TPR) and False Positive Rate (FPR) of this binary classification, we calculate the AUROC that is used as an evaluation metric.

\begin{table*}[t] 
  \caption{Comparison of AUROC to out-of-distribution data (SVHN, LSUN, ImageNet, and CIFAR100) for models trained CIFAR10 on ResNet18 by each data augmentation. The best values are shown in bold. }
  \begin{center}
  \begin{tabular}{c|c|cccc}\hline\hline
    method & Test acc.(\%) & SVHN & LSUN & ImageNet & CIFAR100 \\ \hline
    baseline & 93.50 & 89.22 & 88.61 & 82.68 & 84.88 \\
    CutMix~\cite{CutMix} & 95.00 & 83.74 & 87.26 & 79.24 & 83.18 \\
    mixup~\cite{Mixup} & \textbf{95.31} & 82.91 & 87.41 & 76.63 & 78.09 \\
    APR~\cite{Amplitude} & 95.21 & 98.13 & 92.94 & 84.46 & 88.45 \\
    RFC (proposed) & 94.07 & 98.15 & 91.03 & 83.04 & 85.83 \\
    RFC (proposed) + APR & 94.71 & \textbf{98.59} & \textbf{93.82} & \textbf{85.17} & \textbf{89.02}\\\hline
  \end{tabular}
  \end{center}
  \label{tab:resnet_10}
\end{table*}

\subsection{Results of out-of-distribution detection}
\label{ssec:results}
Table~\ref{tab:resnet_10} shows the accuracy of each data augmentation method and the AUROC values when the models are trained using ResNet18 and CIFAR10 as the training dataset. The accuracy of the data augmentation methods for the spatial domain, such as mixup~\cite{Mixup} and CutMix~\cite{CutMix}, is over 95\%, while the baseline accuracy is 93.50\%. However, the AUROC, which indicates the robustness to out-of-distribution data, drops for all out-of-distribution datasets, especially for SVHN, to around 83\%, which is about 6\% inferior to the baseline (89.22\%). In comparison, the AUROC of our RFC is better than the baseline for all the out-of-distribution datasets, especially for SVHN, with 98.15\%, an improvement of nearly 9\%. In addition, RFC+APR~\cite{Amplitude} yeilds the best AUROC for all the out-of-distribution datasets. The advantage of our proposed RFC is that, as demonstrated in this experiment, it can be combined with other data augmentation methods.

\subsection{Accuracy for CIFAR10-C}
\label{ssec:corrupt}
In this section, we investigate the accuracy of each model in the CIFAR10-C dataset~\cite{CIFAR10-C}. It consists of the CIFAR10 dataset plus 19 types of corruptions in five levels. The accuracy of each model for each corruptions (gaussian noise, gaussian blur, fog, and contrast) of level five intensities is shown in Table~\ref{tab:cor}. The column of averages in the table shows the average accuracy of each model for all 19 types of corruptions in CIFAR10-C. From average accuracy of the table, we can see that RFC+APR is much more robust to corruptions with the accuracy of 75.86\% than the baseline (57.52\%). Although APR is better for fog and contrast with a small margin, RFC+APR is the best in terms of the average performance.
\begin{table*}[t]
  \caption{Comparison of the accuracy of each data augmentation to CIFAR10-C (\%). The best values are shown in bold. }
  \begin{center}
  \begin{tabular}{c|c|ccccc}\hline\hline
    method & Test acc. & \begin{tabular}{c}gaussian\\noise\end{tabular} & \begin{tabular}{c}gaussian\\blur\end{tabular} & fog & contrast & average \\ \hline
    baseline & 93.50 & 27.53 & 31.89 & 73.42 & 48.65 & 57.52 \\
    CutMix~\cite{CutMix} & 95.00 & 30.39 & 22.98 & 76.78 & 67.72 & 59.05 \\
    mixup~\cite{Mixup} & \textbf{95.31} & 41.42 & 50.64 & 79.33 & 72.18 & 68.10 \\
    APR~\cite{Amplitude} & 95.21 & 44.24 & 85.65 & \textbf{90.67} & \textbf{74.78} & 73.27 \\
    RFC (proposed) & 94.07 & 16.94 & 27.27 & 81.34 & 50.59 & 51.71 \\
    RFC (proposed) + APR & 94.71 & \textbf{51.56} & \textbf{86.48} & 89.97 & 73.54 & \textbf{75.86} \\\hline
  \end{tabular}
  \end{center}
  \label{tab:cor} 
\end{table*}

\subsection{Investigation on how the models utilize the frequency components}
\label{ssec:freq}
\begin{table*}[t]
  \caption{Accuracy of the model trained with each data augmentation on various test data for $r=4$ (\%). Values in parentheses are for $r=8$. Low, High means low-frequency and high-frequency respectively. P means using only phase component. }
  \begin{center}
  \begin{tabular}{c|c|ccccc}\hline\hline
    method & Original & Low & High & Low-P & High-P & Phase only \\ \hline
    baseline & 93.50 & 15.40~(31.95) & 10.01~(10.00) & 12.26~(17.47) & 10.00~(10.00) & 73.71 \\
    CutMix~\cite{CutMix} & 95.00 & 13.37~(12.91) & 9.35~(7.49) & 11.27~(9.56) & 10.06~(9.98) & 70.24 \\
    mixup~\cite{Mixup} & 95.31 & 14.45~(45.59) & 9.89~(9.30) & 11.42~(23.17) & 10.00~(9.98) & 79.61 \\
    APR~\cite{Amplitude} & 95.21 & 26.71~(81.07) & 88.77~(68.75) & 20.34~(72.14) & 80.41~(48.65) & 92.61 \\
    RFC (proposed) & 94.07 & 10.01~(17.33) & 89.47~(65.71) & 10.00~(11.54) & 64.24~(26.46) & 74.25 \\
    RFC (proposed) + APR & 94.71 & 48.43~(85.19) & 87.75~(80.15) & 37.60~(74.36) & 80.09~(66.29) & 91.47 \\\hline
  \end{tabular}
  \end{center}
  \label{tab:aug_test} 
\end{table*}
We investigate how differently each model handles frequency components. For this purpose, we use ResNet18 trained with each data augmentation in CIFAR10 and examine the accuracy when the low and high-frequency components of the images are separately input to the model.

Here, we describe how to generate the image with only the phase component. Let $\mathcal{D}_t$ be test datasets and $\bm{x}$ is included in $\mathcal{D}_t$. The frequency component $\mathcal{F}(\bm{x})$ of $\bm{x}$ is calculated using its amplitude $\mathcal{A}_x$ and phase $\mathcal{P}_x$ as shown below:
\begin{eqnarray}
    \mathcal{F}(\bm{x}) = \mathcal{A}_x \otimes e^{i\cdot\mathcal{P}_x}.
\end{eqnarray}
Using this, we can calculate the average amplitude $\mathcal{A}_m$ as
\begin{eqnarray}
    \mathcal{A}_m = \frac{1}{\left| \mathcal{D}_t \right|}\sum_{x \in \mathcal{D}_t} \mathcal{A}_x .
\end{eqnarray}
For each image, the image with only the phase component $\bm{x}^p$ is calculated as:
\begin{eqnarray}
    \bm{x}^p = \mathcal{F}^{-1}(\mathcal{A}_m \otimes e^{i \cdot \mathcal{P}_x}).
\end{eqnarray}
Using the mask matrices $M_l$ and $M_h$ defined in Eq.~(\ref{eq:mask}), the phase-only images of the low-frequency and high-frequency components, $\bm{x}^p_l$ and $\bm{x}^p_h$, are computed as:
\begin{eqnarray}
    \bm{x}^p_l = \mathcal{F}^{-1}(\mathcal{A}_m \otimes M_l \otimes e^{i \cdot \mathcal{P}_x}),\\
    \bm{x}^p_h = \mathcal{F}^{-1}(\mathcal{A}_m \otimes M_h \otimes e^{i \cdot \mathcal{P}_x}).
\end{eqnarray}

The accuracies of the models trained on each data augmentation for these images when $r=4,8$ are shown in Table~\ref{tab:aug_test}.
From Table~\ref{tab:aug_test}, when $r=4$, the accuracy of the baseline model is $10\%$ for high-frequency and high-frequency of phase only components, which is equivalent to a random classifier since CIFAR10 is a 10-class classification. And the baseline model is close to a random classifier for low-frequency components by yielding the accuracy of 15.40\%. Similarly, the performance of mixup is the same as that of the random classifier regardless of the presence or absence of amplitude components in both low and high-frequency components. APR, on the other hand, is slightly more accurate than the baseline by achieving the accuracy about $27$\% for images with only low-frequency components and about $89$\% for high-frequency components. Furthermore, even for the low-frequency and high-frequency components with only phase, the accuracy does not decrease significantly, and it is considered that a large percentage of the judgment is made based on the phase of the image. RFC has the same tendency to the high-frequency component as APR, but is the same as the random classifier when the low-frequency component is used. However, in RFC+APR, the accuracy using the low-frequency component is greatly increased.

Next, we compare the accuracy of each model when $r=8$. Compared to $r=4$, the amount of information in the low-frequency component is larger and that in the high-frequency component is smaller. Therefore, from Table~\ref{tab:aug_test}, the overall trend is that the accuracy  using the low-frequency component is higher and vice versa. For APR and RFC+APR, the accuracy using the low-frequency component increases considerably to more than $80$\%, and the accuracy using the phase of the low-frequency component alone is more than $70$\%. The decline of accuracies of APR and RFC+APR is less than that of baseline (31.95\% to 17.47\%) or mixup (45.59\% to 23.17\%). This indicates that the information of the low-frequency phase is utilized more than that of amplitude. Furthermore, since the accuracy of the high-frequency component of RFC+APR is higher than that of APR by more than $10$\% and the accuracy of the low-frequency component is also higher in RFC+APR, the difference between APR and RFC+APR can be explained as follows: RFC+APR uses the high-frequency component above $r=8$.

Here, let us discuss the relationship between the robustness to out-of-distribution data and the accuracy of low and high-frequency components. The robustness to  out-of-distribution data is generally in the order of RFC+APR $>$ APR $>$ RFC $>$ baseline $>$ CutMix $>$ mixup. Unlike the other models, the baseline and mixup models, which are less robust to out-of-distribution data, are equivalent to random classifiers in terms of accuracy for images with high-frequency components, while the RFC, APR, and RFC+APR models, which are more robust to out-of-distribution data, use more high-frequency components with $r=8$ or higher than the other models. Furthermore, RFC+APR, which has the highest performance, especially utilizes more high-frequency components than the other models. Therefore, it is inferred that the robustness to out-of-distribution data depends on whether the model uses high-frequency components or not. Also, RFC+APR and APR have high accuracy for low-frequency and phase only images, and both models have both accuracy and robustness to out-of-distribution data. Thus, it is expected that the low-frequency or phase is necessary to improve both accuracy and robustness to out-of-distribution data.
In addition, the fact that the RFC utilizes more high-frequency components explains the accuracy of the Section~\ref{ssec:corrupt} with respect to CIFAR10-C. RFC is considerably less accurate against gaussian noise and gaussian blur than the baseline model. Considering the corruptions, low-frequency component of gaussian noise is not so different from the original image, but the high-frequency component deviates greatly from the original image due to noise. Gaussian blur, which is equivalent to a low-pass filter in terms of its operation on the image, is an operation in which the high-frequency components required by RFC are lost. Therefore, for image corruptions such as noise and blur, the high-frequency components are significantly changed or lost, and the accuracy of RFC is supposed to be greatly reduced. On the other hand, noises such as fog and contrast are regarded as operations in which a uniform change in the entire frequency range is applied, and as a result, RFC is expected to be as accurate as baseline in such corruptions. 

\section{Conclusions}
\label{sec:conclusion}
In this paper, we proposed a frequency-based data augmentation that can enhance the robustness to out-of-distribution data. Furthermore, we experimentally showed that robust models mainly use high-frequency of images. It was also suggested that models that also use low-frequency or phase components are also more robust to corrupted data.
Future research may include investigating the robustness to adversarial examples and further frequency-based data augmentation.



\vfill\pagebreak

\bibliographystyle{IEEEbib}
\bibliography{refs}

\end{document}